\documentclass[journal]{IEEEtran}
\usepackage{cite}
\usepackage{multirow}
\usepackage{graphicx}
\usepackage{amssymb,amsmath,bm}
\usepackage{textcomp}
\usepackage[utf8]{inputenc}
\usepackage[subnum]{cases}

\begin{document}

\title{Unsupervised Discovery of Structured Acoustic Tokens with Applications to Spoken Term Detection}
\author{Cheng-Tao~Chung and Lin-Shan~Lee,~\IEEEmembership{Fellow,~IEEE}}
\maketitle

\begin{abstract}

In this paper, we compare two paradigms for unsupervised discovery of structured acoustic tokens directly from speech corpora without any human annotation. 
The Multi-granular Paradigm seeks to capture all available information in the corpora with multiple sets of tokens for different model granularities. 
The Hierarchical Paradigm attempts to jointly learn several levels of signal representations in a hierarchical structure.
%
%
%
The two paradigms are unified within a theoretical framework in this paper. 
%
%
%
%
%
%
Query-by-Example Spoken Term Detection (QbE-STD) experiments on the QUESST dataset of MediaEval 2015 verifies the competitiveness of the acoustic tokens.
The Enhanced Relevance Score (ERS) proposed in this work improves both paradigms for the task of QbE-STD.
We also list results on the ABX evaluation task of the Zero Resource Challenge 2015 for comparison of the Paradigms.
\end{abstract}


\section{Introduction}\label{sec:intro}

\IEEEPARstart{C}{urrent} speech recognition technology is dominated by supervised learning paradigms that rely on massive quantities of human generated linguistic labels. 
Acquiring such human labeled audio is expensive, and it would be hard to scale such efforts with the ever-growing quantity of audio content over the Internet. 
Generating such labeled corpora also requires the knowledge about the language, including its phoneme set and pronunciation lexicon. 
This becomes even more difficult when multiple languages are mixed in the same corpora.
This is why researchers have begun to investigate semi-supervised and unsupervised paradigms that attempt to train acoustic models with limited labeled audio in the low-resource scenario \cite{gales2014speech}, and no labeled audio in the zero-resource scenario \cite{chan2011unsupervised,huijbregts2011unsupervised,siu2014unsupervised,lee2012nonparametric,novotney2009unsupervised,wei2017personalized,kamper2017segmental}.
In this paper we focus our discussion on the unsupervised paradigm where the machine has to learn token-like representations in the speech signal directly from the unannotated corpora.

The two most popular forms of speech signal representation are either a sequence of real-valued frame-level feature vectors (like Mel-Frequency Cepstral Coefficients (MFCC) or spectrogram), or a sequence of discrete tokens (like words or phonemes). 
%
%
Likewise, the works on unsupervised speech technologies extracted either frame-level features \cite{szoke2015copingwith,leung2016toward,chen2016unsupervised,yang2014intrinsic,wang2015acoustic,renshaw2015comparison,zhang2013unsupervised,huang2015rapid} or discrete tokens \cite{park2008unsupervised,jansen2012indexing,jansen2011towards,gish2009unsupervised,chung2013unsupervised,chung2014unsupervised,chung2015enhancing,li2013towards} out of an unlabeled corpus.
To learn unsupervised frame-level representations, Zhang and Glass \cite{zhang2010towards} used posteriorgram features from  unsupervised GMM universal background model (UBM), Chen et al. \cite{chen2015parallel} used posteriorgrams from a
non-parameteric infinite GMM, Kamper et al. \cite{kamperunsupervised} proposed the correspondence autoencoder (cAE): an AE-like deep NN that incorporates top-down constraints by using aligned frames from discovered words as input-output pairs, and we extracted Bottleneck Features (BNF) trained from DNN trained with unsupervised HMMs as targets in our previous work \cite{chung2015iterative}.
To learn discrete tokens, Siu et al. \cite{siu2014unsupervised} used iterative re-estimation and unsupervised decoding procedure of traditional HMMs,  Lee and Glass used the non-parameteric Bayesian HMM \cite{lee2012nonparametric}, Kamper et al. \cite{kamper2017segmental} used embedded segmental K-means models, and we used traditional HMMs with additional constraints in our previous works \cite{chung2013unsupervised,chung2014unsupervised,chung2015enhancing}. 

In this work\footnote{This work was sponsored by the Ministry of Science and Technology, R.O.C.}, we organize our previous contributions  on discrete token learning with unsupervised HMMs, into two paradigms for HMM training.
The Multi-granular Paradigm used the multiple acoustic token sets scattered over a granularity space \cite{chung2014unsupervised,chung2015enhancing}, and can be applied to semi-supervised speaker adaptation \cite{wei2017personalized}.
The Hierarchical Paradigm used the two-level word-like and subword-like tokens \cite{chung2013unsupervised,chung2015iterative}, and can be used in query expansion of semantic retrieval systems for spoken documents \cite{lee2013enhancing,li2013towards}.
Both of them produced structured acoustic tokens, rather than a single set of acoustic tokens. 
We unify the paradigms with one framework, and compare the two paradigms on the same experiments.

Query-by-Example Spoken Term Detection (QbE-STD) was  chosen as the example application to compare tokens trained with the two paradigms on.
%
%
QbE-STD refers to the task of finding all occurrences of the input spoken query from a large target audio corpus. 
Most QbE-STD approaches were based on automatic speech recognition (ASR), transforming speech into words or subwords for token matching \cite{miller2007rapid,mamou2007vocabulary,wallace2007phonetic,pan2010performance}, with performance relying heavily on the ASR accuracy \cite{saraclar2004lattice}. 
This implies annotated training corpora properly matched to the spoken content are necessary. 
Because both the input query and the target corpus are spoken, it is possible to directly match the spoken query to the target corpus without knowing the written form of either the query and the target corpus, bypassing the need for supervised speech recognition.
This is especially attractive for languages with limited annotated data \cite{boves2009resources,kumar2007wwtw} or spoken content with unknown languages. 
%
%
%
%
There are in general two approaches to compute the distance between a spoken query and a spoken document (an utterance in the target audio corpus):  comparing the acoustic feature sequences directly, or transcribing audio files into sequences of acoustic tokens, then comparing the transcribed token sequences.
%
%
The former approach is easily affected by speaker mismatch and varying acoustic conditions. The high computation cost also makes it hard to scale for larger spoken corpora
%
%
%
\cite{chan2011unsupervised,huijbregts2011unsupervised,carlin2011rapid,zhang2009unsupervised,wang2012acoustic,zhang2011piecewise,zhang2012fast,jansen2012indexing}.
%
%
%
The latter approach smoothens signal variations by token models and has the advantage of much lower on-line computation requirement when the target corpus is large. 
The methods proposed in this paper belongs to the latter approach.
The rest of the work is organized as follows.
The Multi-granular Paradigm is explained in Section \ref{sec:mult_par} and the Hierarchical Paradigm is explained in Section \ref{sec:hier_par}.
We unify the two paradigms with theoretical analysis in Section \ref{sec:unification}, and reason that the Hierarchical Paradigm can be reduced to the Multi-granular Paradigm when certain criteria are met.
The Query-by-Example Spoken Term Detection (QbE-STD) using the tokens is explained in Section \ref{sec:distance}.
The setting of either training the tokens on the documents or the queries for the QbE-STD experiments are explained in Section \ref{sec:setting}.
%
%
%
%
The QbE-STD experiments on the QUESST dataset  \cite{szoke2015query} of MediaEval 2015 in Section \ref{sec:std} supports our framework and verified the competitiveness of the acoustic tokens.
%
%
%
Additional results on the Zero Resource Challenge ABX evaluation task \cite{versteegh2015zero} also resented in Section \ref{sec:abx}.
Finally, we explain how to choose which Paradigm to use in Section \ref{sec:support} and give our concluding remarks in Section \ref{sec:conclusion}.

\section{The Paradigms}\label{sec:paradigm}
Given an unlabeled speech corpus, it is straightforward to discover acoustic tokens using unsupervised HMMs with a chosen configuration.
%
We assume the number of states $m$ in each HMM, and the total number of distinct acoustic tokens $n$ is the same for all HMMs in the HMM set we consider during initialization, with $\psi=(m,n)$ used to represent the model configuration. 
%
%
In each iteration $i$, we then train the HMM parameters $\theta_i^{mn}$ with the label sequences $W_{i-1}$ obtained in the previous iteration as in Eq. (\ref{eq:2}) and decode the label sequences $W_{i}$ with the obtained parameters $\theta_i^{mn}$ as in Eq. (\ref{eq:3}) \cite{gish2009unsupervised}. 
\begin{eqnarray}
\theta_{i}^{mn} &=& \arg \max_{\substack{\theta^{mn}}} P(X|\theta^{mn},W_{i-1}),   \label{eq:2} \\
W_i &=& \arg \max_{\substack{W}} P(X|\theta_i^{mn} ,W),             \label{eq:3}
\end{eqnarray}
where $X$ is the acoustic vector sequence for the whole corpus being considered.
Eq. \eqref{eq:2} is the maximum likelihood training of HMMs, and the only difference here is that we train on the label sequences $W_{i-1}$ obtained in the previous iteration rather than ground truth labels. 
Eq. \eqref{eq:3} is the Viterbi decoding based on the model set $\theta_i^{mn}$.
This method to train HMMs was introduced by Gish et al. \cite{gish2009unsupervised}.
To find a better initial label, we used the method explained in \cite{chung2013unsupervised} to obtain $W_0$.
To model more complex structures in speech, we proposed two natural ways to extend this method: (1) Use multiple sets of HMMs with different configurations \cite{chung2014unsupervised} which we call the Multi-granular Paradigm in this work, (2) Combine the HMMs into longer tokens to construct language structures \cite{chung2013unsupervised} which we call the Hierarchical Paradigm in this work.
The two Paradigms are explained below.

\subsection{The Multi-granular Paradigm}\label{sec:mult_par}
In this paradigm, we take into consideration the cases where several intrinsic acoustic representations exist in the spoken corpus.
Acoustic units with different lengths such as phonemes, syllables, words, and phrases have different temporal granularity.
Different phonetic clusters such as speaker-independent phonemes, gender-dependent phonemes, speaker-dependent phonemes have different phonetic granularity. 
We wish to develop a set of acoustic token HMMs for every different granularity configuration (temporal and phonetic) on the same corpus.
With multiple sets of granularity configurations, we can have multiple sets of acoustic token HMMs trained on the same corpus.
We call the level of representation corresponding to a granularity configuration a layer in the discussion below.
%


\begin{figure}[t]
\centerline{\includegraphics[width=0.45\textwidth]{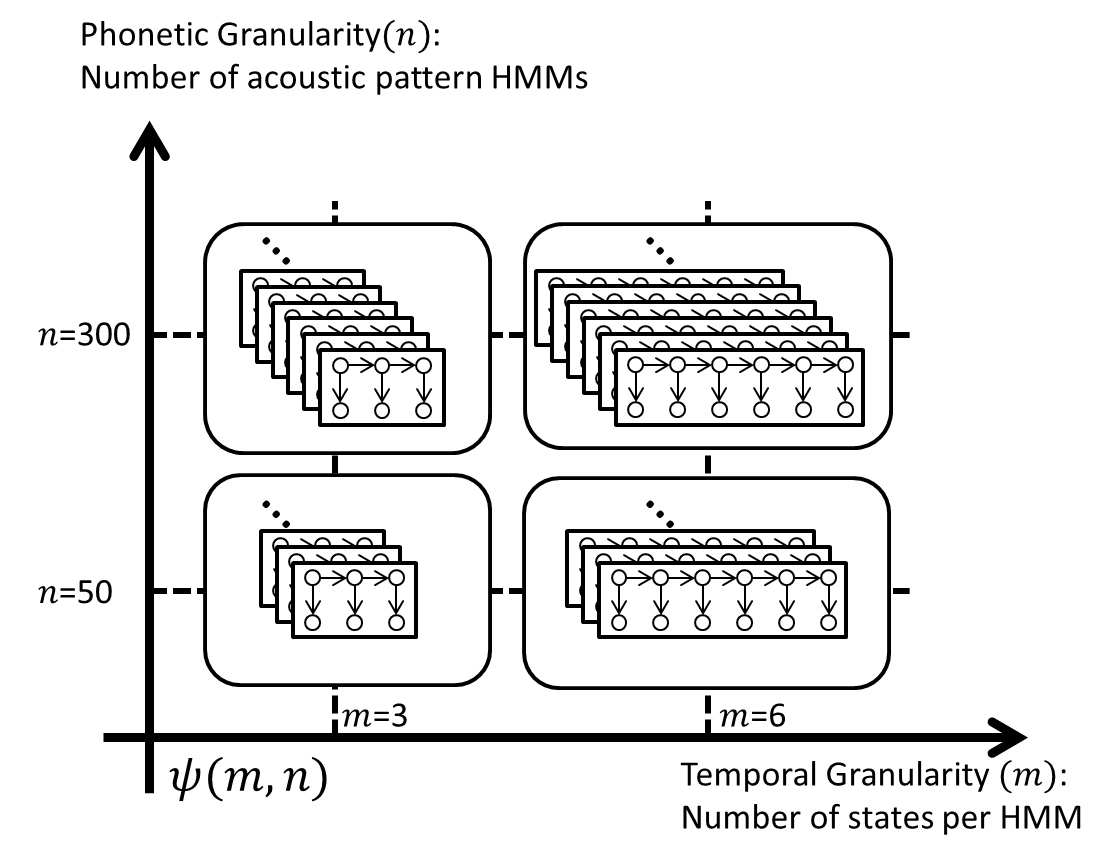}}
\caption{Model granularity space for acoustic token configurations}\label{fig:2dcube}
\end{figure}

Here each layer of acoustic token is defined by a set of temporal and phonetic granularity parameters.
%
Different layers of tokens are discovered independently of each other in the Multi-granular Paradigm.
%
%
The transcription of a signal decoded with these tokens can be considered as a temporal segmentation of the signal, so the HMM length (or number of states in each HMM) $m$ represents the temporal granularity. 
The set of all distinct acoustic tokens can be considered as a segmentation of the phonetic space, so the total number $n$ of distinct acoustic tokens represents the phonetic granularity. 
This gives a two-dimensional representation of the acoustic token configurations in terms of temporal and phonetic granularities as in Fig. \ref{fig:2dcube}. 
Any point $\psi=(m,n)$ in this two-dimensional space in Fig. \ref{fig:2dcube} corresponds to an acoustic token configuration. 
Note that there can be a third dimension, the acoustic granularity which is the number of Gaussians in each HMM state, but the effect of that dimension has been shown to be negligible in the experiments, thus we simply set the number of Gaussians in each state to be 4. 
Although the selection of the hyperparameters can be arbitrary in this two-dimensional space in principle, here we simply select $M$ temporal granularities and $N$ phonetic granularities, forming a two-dimensional array of $M \times N$ hyperparameter sets in the granularity space and $M \times N$ sets of acoustic tokens. We denote the collection of selected granularities to be $G$, as shown in Eq. (\ref{eq:G}).
\begin{equation}
G=\{(m_j, n_k)| 1\leq j \leq M, 1\leq k \leq N\} \label{eq:G}.
\end{equation}

Note that the optimal granularities $(m,n)$ can be dependent on the spoken corpus and there is no emphasis on discovering acoustic tokens close to linguistically meaningful units for the Multi-granular Paradigm. 
%
Tokens at each granularity may not correspond to any of these linguistically meaningful units, but with multiple sets of tokens, most of the structures and characteristics of the corpus can be captured if the granularities are evenly distributed over the granularity space.
%
%

\subsection{The Hierarchical Paradigm}\label{sec:hier_par}
The Hierarchical Paradigm, on the other hand, automatically learns the word-like and subword-like tokens, as well as all relevant knowledge for this set of tokens such as the lexicon and N-gram language model referred to as the linguistic structure for the spoken language of the given corpus.
The goal is to find the parameters for the linguistic structure, and the word-like token label sequences $W$ on the observed acoustic feature vector sequences $X$ for the corpus considered which was discussed in \cite{chung2013unsupervised}. 
In this work we leave out the language models in the discussion to compare the Paradigms, so the parameter set  $\theta^{mn}_{uv}=\{\theta^{mn},\theta_{uv}\}$ includes two parts: $\theta^{mn}$ for acoustic HMMs of subword-like tokens, $\theta_{uv}$ for the lexicon of word-like tokens in terms of subword-like token sequences.
$u$ denotes the number of states of the longest word-like tokens in the lexicon, and $v$ denotes the number of words in the lexicon.
The granularity parameters $(m,n)$ are chosen, $u$ is set to an integer multiple of $m$, and $v$ is inferred.
Contrary to the Multi-granular Paradigm where acoustic tokens with multiple granularities $(m_j, n_k)$ are trained independently, in the Hierarchical Paradigm the subword-like tokens and word-like tokens are like two correlated granularities $(m,n)$  and $(u,v)$ that are jointly trained.
%
%
The iterations in Eq. (\ref{eq:2}) and Eq. (\ref{eq:3}) can be further segmented into several cascaded stages that uses slightly different objectives for acoustic, linguistic and lexical optimization. 
Two of these stages are summarized below, and more details of such stages can be found in our previous work \cite{chung2013unsupervised}.
%
%
When the difference between $W_{i-1}$ and $W_{i}$ becomes insignificant, the process then advances to the next stage. 
%
%
%
The basic idea behind the procedure of having multiple stages is to gradually construct and update the parameters from subword-like tokens to word-like tokens. 
This prevents the parameters from being caught in local optimal situations which often happen when too many parameters are optimized at the same time.
The general flow of the training procedure is as follows.

In the acoustic optimization stage, $\theta_{uv}$ is fixed and the HMM parameters $\theta^{mn}$ for the subword-like tokens are trained alone, because these HMMs are the primary building blocks of the whole linguistic structure and reliable estimate for their parameters is the key. 
In each iteration, the acoustic model set $\theta^{mn}$ are the HMMs trained from the corpus based on $W_i$ with the ML criterion as in Eq. \eqref{eq:2}.
The lexicon $\theta_{uv}$ is derived by collecting all word-like tokens appearing in $W_i$ with counts exceeding a threshold.
Free word decoding is then performed on the whole corpus $X$ based on $\theta^{mn}$ and $\theta_{uv}$, producing an updated label $W_{i+1}$.
When $W_i$ is updated to $W_{i+1}$, not only the HMM parameters of $\theta^{mn}$ and HMM segmentation boundaries are updated, but the vocabulary size $v$ of $\theta_{uv}$ may shrink when the counts of some word-like tokens become small enough.

In the lexical optimization stage, we then break the word-like tokens into subword-like tokens and reconstruct better word-like tokens in the lexicon.
In each iteration, we reconstruct new word-like tokens by breaking the existing word-like tokens in $\theta_{uv}$ into subword-like tokens, and then reconstructing new word-like tokens based on their occurrence in $W_i$. 
Those segments of several consecutive subword-like tokens appearing frequent enough and with high enough right and left context variation are taken as new word-like tokens. 
This can be realized by constructing an efficient data structure called PAT-Tree using the labels $W_i$\cite{ong1999updateable}. 
In this way, the lexicon $\theta_{uv}$ can be updated significantly in each iteration. 
With the PAT-Tree, word-like tokens consisting of a single subword-like token are often included into the lexicon.
If a word is not in the lexicon, it is usually represented by a sequence of subword-like tokens.
This updated lexicon $\theta_{uv}$ is then used in free-word decoding to produce the labels $W_{i+1}$. 
The whole process is completed when there is no significant difference between $W_i$ and $W_{i+1}$. 
This gives the automatically discovered linguistic structure $\theta^{mn}_{uv}=\{\theta^{mn},\theta_{uv}\}$.
%
The time alignment for the subword-like tokens are updated in all iterations when the labels $W_i$ are decoded.

Note that the word-like and subword-like tokens here try to mimic the linguistically meaningful words and subword units, but for some arbitrary granularity $(m,n)$, the results may not be linguistically meaningful at all (close to phonemes, words, phrases, etc.). 
However, if the granularities $(m,n)$ are properly chosen such that the discovered subword-like tokens are close to the phonemes of the language, then discovered word-like tokens in the lexicon $\theta_{uv}$ can be really close to the words of the language. 
The point of having word-like tokens and the lexicon
is to provide higher-level context constraints which help produce the subword-like tokens with higher quality.

\section{A unified view of the paradigms}\label{sec:unification}
Although the Hierarchical Paradigm and Multi-granular Paradigm may look different at first glance, in this section we propose a unified view on these two paradigms.
The objective of this section is to offer a guideline for selecting the granularity parameters under the Multi-granular Paradigm so that the two Paradigms can be compared in terms of their complexity.
The word-like tokens under the Hierarchical Paradigm can be considered as tokens of a coarser temporal granularity under the Multi-granular Paradigm.
The tokens of a finer temporal granularity under the Multi-granular Paradigm can be considered as subword-like tokens used to construct the word-like tokens under the Hierarchical Paradigm.
With this approach, we can compare one paradigm to the other easily and characterize the mathematical relation between the two.

Let the total number of possible token sequences on any utterance for a token set with model parameter $\theta$ be denoted as $C(\theta)$.
For an utterance of $T$ frames, we assume that each HMM state occupies the same number of frames $d$.
Let us denote the model parameters of a Multi-granular token set with granularity $(m,n)$ as $\theta^{mn}$.
Under the Multi-granular Paradigm, the length of every HMM is $md$, and the utterance is represented as a token sequence of $T/md$ acoustic tokens.
There are $n$ possible acoustic tokens, so the number of possible token sequences on an utterance of length $T$ is
\begin{equation}
 C(\theta^{mn}) = n^{\frac{T}{md}}. \label{eq:card_mult}
\end{equation}
%


Model parameters for a token set under the Hierarchical Paradigm consist of two parts: 
%
%
Let $\theta^{mn}_{uv}=\{\theta^{mn},\theta_{uv}\}$
, or the subword-like token acoustic model $\theta^{mn}$ of granularity $(m,n)$ and the word-like token lexicon $\theta_{uv}$.
Because a word-like token is composed of one to several subword-like tokens, $u$ is a positive integer multiple of $m$.
%
%
Since the lexicon $\theta_{uv}$ limits the token sequence for any utterance to include only word-like tokens in the lexicon, the possible subword-like token sequences is a subset of the case with only the subword-like acoustic token model $\theta^{mn}$, hence we have
\begin{equation}
C(\theta^{mn}_{uv}) \leq C(\theta^{mn}). \label{eq:card_upper}
\end{equation}
On the other hand, with the Hierarchical Paradigm we represent the utterances as sequences of word-like tokens.
We have a total of $v$ allowed distinct word-like tokens and the maximum length of a word-like token is $u$.
%
%
Since not all word-like tokens are composed of the same number of subword-like tokens in the lexicon, this set of $v$ word-like tokens would have mixed temporal granularity. This means the $v$ word-like tokens would have varying number of states with the longest having $u$ states.
The lower bound of the total number of token sequence representations for this situation is the case when all the $v$ word-like tokens in the lexicon have the same length of $u$ states.
This number is the same as a set of $v$ subword-like tokens each with $u$ states, or $\theta^{uv}$ for the Multi-granular Paradigm, therefore 
\begin{equation}
C(\theta^{uv}) \leq C(\theta^{mn}_{uv}). \label{eq:card_lower}
\end{equation}
%
%
By combining Eq. \eqref{eq:card_upper} and Eq. \eqref{eq:card_lower}, we can get an upper and a lower bound:
\begin{equation}
 C(\theta^{uv}) \leq C(\theta^{mn}_{uv}) \leq  C(\theta^{mn}). \label{eq:card_hier}
\end{equation}
%


%
Eq. \eqref{eq:card_hier} means that we can bound the number of possible token sequence representations for utterances based on a token set $\theta^{mn}_{uv}$ under the Hierarchical Paradigm using two token sets $\theta^{mn}$ and $\theta^{uv}$ under the Multi-granular Paradigm.
This is helpful, because acoustic tokens under the Hierarchical Paradigm can be tricky to train.
With Eq. \eqref{eq:card_hier}, we can achieve similar order of representations with the Multi-granular Paradigm by selecting the granularities within the box $\overline{G}$ on the granularity plane,
\begin{equation}
\overline{G} = \{(\overline{m_i}, \overline{n_j})| m\leq \overline{m_i} \leq u, n\leq \overline{n_j} \leq v\} \label{eq:card_box},
\end{equation}
between the points $(u,v)$ and $(m,n)$.
This also serves as a good guideline for selecting the granularity parameters under the Multi-granular Paradigm when some knowledge about the underlying language for the spoken corpus is known:
the temporal granularities can be selected ranging between those corresponding to the duration of an average phone $m$, and of the longest word $u$; while the phonetic granularities can be selected ranging between the  size of the phoneme inventory $n$ and the size of the vocabulary $v$.

Recall that in the lower bound in Eq. \eqref{eq:card_hier}, $\theta^{uv}$ is actually the granularity parameter for a token set under the Multi-granular Paradigm, so we may rewrite it as $\theta^{m'n'}$.
By substituting $(u,v)$ with $(m',n')$, and using Eq. \eqref{eq:card_mult} with the relation $C(\theta^{m'n'})\leq C(\theta^{mn})$ in \eqref{eq:card_hier}, we get 
\begin{equation}
\frac{\log n'}{m'} \leq \frac{\log n}{m}. \label{eq:card_log}
\end{equation}
The equality in Eq. \eqref{eq:card_log} becomes true when all words in the corresponding lexicon are composed of the same number of states $m'$ and
\begin{equation}
n'=n^\frac{m'}{m}. \label{eq:card_equib}
\end{equation}
When Eq. \eqref{eq:card_equib} holds, the number of possible representations for a token set $\theta^{mn}_{m'n'}$ under the Hierarchical Paradigm actually is the same as that of a token set $\theta^{mn}$ or $\theta^{m'n'}$ under the Multi-granular Paradigm (lower bound is equal to the upper bound). 
In other words, the two-level Hierarchical Paradigm is reduced to the one-level Multi-granular Paradigm.
The simplest situation for this to happen would be when every subword-like token is a word-like token in the lexicon ($m'=m$,$n'=n$).
Another case is when all word-like tokens are composed of exactly $k$ subword-tokens, and every possible combination of $k$ subword-like tokens corresponds to a word-like token in the lexicon. ($m'=km$,$n'=n^k$).
An example would be when $m=3$, $n=50$ and $m'=6$, $n'=2500$.
%
This explains why the number of possible representations for utterances by token sequences can be used to develop a unified view to analyze the two paradigms.
%

Note that exact same results in this section can be derived if we used the number of bits required to store the decoded acoustic tokens in our discussion.
For example, we can derive Eq. \eqref{eq:card_equib} directly by equating the token storage  requirements for token sequences in Table \ref{tab:summary} on two different granularities $(m,n)$ and $(m',n')$.
%
%
%
%
%

\section{Spoken Term Detection Using Discovered Acoustic Tokens}\label{sec:distance}

There can be various applications for the acoustic tokens presented here, while Query-by-Example Spoken Term Detection (QbE-STD) is a good example. 
In this section we summarize the ways to perform QbE-STD \cite{chung2014unsupervised} using the acoustic tokens.
Note that due to the nature of unsupervised learning, tokens with high occurrence in the corpus would have better representations, making the system perform relatively worse for queries with low occurrence.

\subsection{Off-line Phase: Token Pairwise-Distance}\label{ssec:std_off}
Let \{$t_i| i=1,2,3,..,n$\} denote a set of the $n$ acoustic tokens discovered here.
For the Multi-granular Paradigm, $n$ is the hyperparameter for phonetic granularity in the parameter set $\psi=(m,n)$ for each set of acoustic tokens.
For the Hierarchical Paradigm $n$ is the total number of subword-like tokens and $t_i$ is a subword-like token. 
We first construct a distance matrix $S$ of size $n \times n$ off-line for these $n$ tokens, for which the element $S(i,j)$ is the distance between any two token HMMs $t_i$ and $t_j$ in the set,
\begin{equation}
S(i, j) =\mbox{KL}(t_i, t_j). \label{eq:soft}
\end{equation}
The KL-divergence $\mbox{KL}(i,j)$ between two token HMMs in Eq. (\ref{eq:soft}) can be defined as the symmetric KL-divergence between each corresponding state in $t_i$ and $t_j$ based on the variational approximation \cite{hershey2007approximating} summed over the states. 
It is also possible to perform a state-level Dynamic Time Warping (DTW) between the two state sequence in $t_i$ and $t_j$ (i.e. one state in HMM $t_i$ can be matched to several states in HMM $t_j$ and vise versa), then sum over the optimal path.
This $S(i,j)$ is constructed for each token set $\psi=(m,n)$.

\begin{figure}[b]
\centerline{\includegraphics[width=0.45\textwidth]{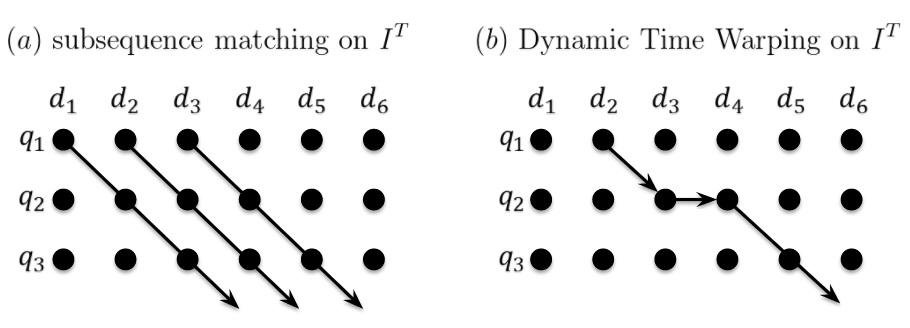}}
\caption{The transpose of the matching matrix $I$}\label{fig:quant}
\end{figure}

\subsection{On-line Phase: Relevance Score Evaluation}
In the on-line phase, we first perform the following for each spoken query $q$ and each spoken document (an utterance) $d$ in the target corpus. 
This is done for each token set $\psi=(m,n)$ in the Multi-granular Paradigm.
%
%
Assume a document $d$ is decoded into a sequence of $D$ acoustic tokens with indices $(d_1, d_2, ..., d_D)$ and the query $q$ into a sequence of $Q$ tokens with indices $(q_1, q_2, ..., q_Q)$.
We thus construct a matching matrix $I$ of size $D \times Q$ for every document-query pair, in which each entry $(i,j)$ is the relevance score between acoustic tokens with indices $d_i$ and $q_j$ as in Eq. (\ref{eq:topk}) and shown in Fig. \ref{fig:quant}(a) for a simple example of $Q=3$ and $D=6$, where $S(i,j)$ is defined in Eq. (\ref{eq:soft}),
\begin{equation}
I(i, j)  = -S(d_i, q_j).  \label{eq:topk}
\end{equation}
The above only considers the one-best token sequences $(t_{d_1}, t_{d_2}, ..., t_{d_D})$ and $(t_{q_1}, t_{q_2} ..., t_{q_Q})$ decoded from $d$ and $q$.
It is possible to consider the N-best token sequences by representing the N-best token sequences as sequences of posteriorgram features and integrate them in the matrix $I$ as shown in \cite{chung2014unsupervised}. 
However, experiments showed that the extra improvements brought in this way is almost negligible in the Multi-granular Paradigm, probably because in that paradigm the $M \times N$ different token sequences based on the $M \times N$ different token sets can be considered as a huge lattice including many one-best paths which are jointly considered here \cite{chung2014unsupervised}. 

For matching the sub-sequence of $d$ with $q$, we take the summation of the elements in the matrix $I$ in Eq. \eqref{eq:dtw} along the diagonal direction, generating the accumulated distance for all sub-sequences starting at all token positions in $d$ as shown in Fig. \ref{fig:quant}(a). 
The minimum is taken as the Initial Relevance Score (IRS) between document $d$ and query $q$, $\widehat{R}(d,q)$ on the token set $\psi=(m,n)$ as in Eq. \eqref{eq:dtw}.
\begin{equation}
\widehat{R}(d,q) = \max_{\substack{i }}\sum_{j=1}^{Q} I(i+j,j).  \label{eq:dtw}
\end{equation}

In this work we propose that the HMM probabilities for the spoken utterances $d$, $q$ evaluated with respect to each token $t_{d_i}$ and $t_{q_j}$ in the token sequences can also be taken into account to produce the Enhanced Relevance Score (ERS) as shown in Eq. \eqref{eq:dtw_p}.
\begin{equation}
R(d,q) = \max_{\substack{i }}\sum_{j=1}^{Q}[ I(i+j,j)+\log P(t_{d_i},d)+\log P(t_{q_j},q)].  \label{eq:dtw_p}
\end{equation}
%
Here $P(t_{d_i},d)$ and $P(t_{q_j},q)$ are the probabilities for observing the tokens $t_{d_i}$ and $t_{q_j}$ in $d$ and $q$ respectively.
%
%
To turn the probabilities into scores, we take the logarithm of $P(t_{q_j},q)$ and $P(t_{d_i},d)$.
Although we are summing over $j$ in Eq. \eqref{eq:dtw_p}, and the sum over $\log P(t_{q_j},q)$ will be the same for a given query $q$, we still keep the term because $\sum 
\log P(t_{q_j},q)$ will be different across acoustic token sets, which matters when we combine the distances.

It is also possible to consider token-level DTW on the matrix $I$ as shown in Fig. \ref{fig:quant}(b). 
%
%
However, experiments have shown that the extra improvements brought in this way is almost negligible in the Multi-granular Paradigm.
This is probably because in that paradigm the different token sequences based on the different token sets (e.g. including longer /shorter tokens) are jointly considered, so the different time-warped matching and insertion/deletion between $d$ and $q$ is already automatically included \cite{chung2014unsupervised}.
%
For the Multi-granular Paradigm, the multiple relevance scores $R(d,q)$ and $\widehat{R}(d,q)$ in Eq. \eqref{eq:dtw} and Eq. \eqref{eq:dtw_p} obtained with multiple token sets can be normalized across documents and then averaged. 
The averaged scores are then used to rank all the documents for QbE-STD. 

\subsection{The Heuristics behind Token Sequence Representations}\label{sec:prince}
%
The task of QbE-STD tries to define some tractable relevance score $R'(d,q)$ between a spoken document $d$ and a spoken query $q$ to approximate the oracle relevance score $R_o(d,q)$ which is unknown.
The role of the unsupervised token sets is to map the two spoken utterances $d$ and $q$ to an intermediate representation where some tractable relevance score $R'(d,q)$  calculation can be performed.
We map $d$ and $q$ to the intermediate representations $F_\theta(d)=(d_1,d_2, ... ,d_D)$ and $F_\theta(q)=(q_1,q_2, ... , q_Q)$ using the token set with parameter $\theta$.
Conceptually, the oracle relevance score $R_o(d,q)$ can be approximated by the Initial Relevance Scores (IRS) evaluated directly from the token sequences,
\begin{equation}
R_o(d,q)\approx  \widehat{R}(d,q) \equiv  R'(F_\theta(d), F_\theta(q)) \label{eq:oracle}
\end{equation}
where $\widehat{R}(d,q)$ is defined in Eq. \eqref{eq:dtw} and $R'(d,q)$ is evaluated with $F_\theta(d)$ and $F_\theta(q)$.
Because the token sequences $F_\theta(d)$ and $F_\theta(d)$ are imperfect representations of $d$ and $q$, $R'(F_\theta(d),F_\theta(q))$ can be far from $R_o(d,q)$.
%
%
The HMM probabilities $P(t_{d_i},d)$ and $P(t_{q_j},q)$ in Eq. \eqref{eq:dtw_p} describe the quality of these representations. 
When these probabilities are high, the approximation in Eq. \eqref{eq:oracle} is closer to the $R_o(d,q)$.
The two terms $\log P(t_{d_i},d)$ and $\log P(t_{q_i},q)$ scores the quality of the representation when we approximate the utterances $d$ and $q$ with the token sequences $F_\theta(d)$ and $F_\theta(q)$.
In Eq. \eqref{eq:dtw_p} we simply add the quality scores to Eq. \eqref{eq:dtw} to produce the Enhanced Relevance Scores (ERS).
\begin{equation}
R_o(d,q) \approx \widehat{R}(d, q) + \mbox{Quality of $F_\theta(q)$, $F_\theta(q)$} 
\end{equation}
In the experiments below we will show that Eq. \eqref{eq:dtw_p} worked better than Eq. \eqref{eq:dtw} in all cases.

\section{Testing Scenarios for the Experiments}\label{sec:setting}
We wish to use QbE-STD as the example application to test the token sets discussed here.
The task of QbE-STD involves two sets of utterances: the spoken queries $Q$ and spoken documents $D$.
If we ignore the structural differences, they are simply two sets of spoken utterances.

In our previous works \cite{chung2013unsupervised,chung2014unsupervised,chung2015iterative}, it was assumed that the spoken queries $Q$ were not accessible during the off-line phase.
That means we assume only one spoken query was given during the on-line phase for every query search, so the retrieved results for every query was independent of each other.
However because the documents $D$ and $Q$ are simply two sets of utterances, in this work we also try to investigate another testing scenario where the situation is reversed.
This means the entire set of spoken queries $Q$ is available during the off-line phase, but the spoken documents $d\in D$ are given one-by-one during the on-line phase.
We further discuss these two scenarios below, both of which will be tested with the token sets proposed.

\subsection{Document Tokens}\label{sec:use_doc}
In this scenario the spoken query set $Q$ is available only during testing, while the whole spoken document set $D$ is available during training of the token sets.
Because the token sets are trained with $D$, tokens trained under this scenario is referred to as document tokens.
%
%
This is the scenario for most cases of STD including our earlier work \cite{chung2013unsupervised,chung2014unsupervised,chung2015iterative}, where $D$ is both the archive containing the spoken documents which we wish to retrieve from, and the archive used to train the unsupervised tokens.

\subsection{Query Tokens}\label{sec:use_qer}
In this scenario the spoken document set $D$ is available only during testing, while the whole spoken query set $Q$ is available during training of the token sets.
Because the token sets are trained with $Q$, tokens trained under this scenario is referred to as query tokens.
This system has the benefit of being very fast to train since we know that the training time complexity is quadratic in the length of the training utterances (from Table \ref{tab:summary} and Appendix \ref{sec:complexity}), and usually queries are much shorter than documents.
Because most queries are short, the quality of the system depends highly on the number of utterances in $Q$.
%
%
Note that when the number of training queries in $Q$ is small, this testing scenario becomes similar to DTW over raw features.
Since $Q$ is small, each HMM will be fed with less training examples.
In the most extreme case, each HMM will be assigned only one training sequence which it sets as the mean of the states, and builds a small variance around it.
Since it has been shown that HMM decoding is in fact very similar to DTW \cite{juang1984hidden}, when calculating the Gaussian state emission probabilities on HMMs, we are computing the Gaussian kernel between the mean of the Gaussian and the frame-level feature.
The QbE-STD process would be similar to performing DTW of each query $q\in Q$ directly over each spoken document $d\in D$ with a Gaussian kernel as a distance measure over the feature pairs.

A real world example for this scenario would be using personalized devices like mobile phones to search for spoken archives. 
The device has a record of spoken queries, possibly stored locally, but not the spoken documents.
The user then decides to search for the queries of interest in different spoken archives that contain the documents in the cloud. 
Note that the user does not have to label the queries in advance.


\begin{table*}[tbh]
\centering
\caption{STD performance of Multi-granular and Hierarchical Document Tokens trained on the Document Corpus of QUESST 2015. $m$ denotes to the number of HMMs in the corpus, $n$ denotes the number of HMMs, $u$ denotes the number of states of the longest word-like token in the lexicon, $v$ denotes the number of word-like tokens in the lexicon. The best result for each column is shown in bold.}
\label{tab:std_doc}
\begin{tabular}{|c|c|c|c|c|c|c|c|}
\hline
\multirow{2}{*}{Paradigm} & \multirow{2}{*}{(m, n)} & \multirow{2}{*}{(u, v)}&\multicolumn{5}{|c|}{model-dependent minCnxe}\\ \cline{4-8}
& & & \phantom{(T}IRS\phantom{0)} & \phantom{(T}ERS\phantom{0)} & ERS(T1) & ERS(T2) & ERS(T3)\\ \hline
\multirow{7}{*}{Multi-granular}
&(3, 100)    &-           &0.8008  &0.7838                    &0.7820  &0.7967  &0.8128\\                       
&(3, 200)    &-           &0.7985  &0.7842                    &0.7860  &0.7942  &0.8001\\
&(5, 100)    &-           &0.7872  &0.7651                    &0.7722  &0.7886  &0.7732\\
&(5, 200)    &-           &0.7929  &0.7671                    &0.7757  &0.7868  &0.7904\\
&(7, 100)    &-           &0.7938  &0.7672                    &0.7727  &0.7908  &0.7885\\
&(7, 200)    &-           &\textbf{0.7833}  &\textbf{0.7645}  &0.7735  &\textbf{0.7797}  &0.7769\\
&average     &-           &0.7826  &0.7644                    &0.7701  &0.7770  &0.7788\\ \hline
\multirow{7}{*}{Hierarchical}                                           
&(3, 100)    &(6, 3338)   &0.7974  &0.7820                    &0.7845  &0.8003  &0.7966\\
&(3, 200)    &(6, 8571)   &0.7978  &0.7825                    &0.7840  &0.7987  &0.7953\\
&(5, 100)    &(10, 3904)  &0.7914  &0.7663                    &\textbf{0.7713}  &0.7861  &\textbf{0.7613}\\
&(5, 200)    &(10, 9191)  &0.7978  &0.7729                    &0.7780  &0.7895  &0.7729\\
&(7, 100)    &(14, 4119)  &0.7939  &0.7700                    &0.7749  &0.7852  &0.7888\\
&(7, 200)    &(14, 8296)  &0.7859  &0.7660                    &0.7736  &0.7798  &0.7920\\
&average     &-           &0.7840  &0.7644                    &0.7706  &0.7805  &0.7915\\ \hline
\end{tabular}
\end{table*}

\begin{table*}[tbh]
\centering
\caption{STD performance of Multi-granular and Hierarchical Query Tokens trained on the Development Queries of QUESST 2015. $m$ denotes to the number of HMMs in the corpus, $n$ denotes the number of HMMs, $u$ denotes the number of states of the longest word-like token in the lexicon, $v$ denotes the number of word-like tokens in the lexicon. The best result for each column is shown in bold.}
\label{tab:std_qer}
\begin{tabular}{|c|c|c|c|c|c|c|c|}
\hline
\multirow{2}{*}{Paradigm} & \multirow{2}{*}{(m, n)} & \multirow{2}{*}{(u, v)}&\multicolumn{5}{|c|}{model-dependent minCnxe}\\ \cline{4-8}
& & & \phantom{(T}IRS\phantom{0)} & \phantom{(T}ERS\phantom{0)} & ERS(T1) & ERS(T2) & ERS(T3)\\ \hline
\multirow{7}{*}{Multi-granular}
&(3, 100)    &-           &0.7938  &0.7856                    &0.7860  &0.7987  &0.8139\\                              
&(3, 200)    &-           &0.8047  &0.7847                    &0.7794  &0.8048  &0.8145\\
&(5, 100)    &-           &\textbf{0.7854}  &\textbf{0.7804}  &0.7862  &0.7984  &0.8062\\
&(5, 200)    &-           &0.8010  &0.7882                    &0.7813  &0.7984  &0.8117\\
&(7, 100)    &-           &0.7908  &0.7814                    &0.7795  &0.7990  &0.8075\\
&(7, 200)    &-           &0.7966  &0.7911                    &0.7807  &0.8014  &0.8073\\
&average     &-           &0.7828  &0.7735                    &0.7747  &0.7984  &0.8114\\ \hline
\multirow{7}{*}{Hierarchical}                                           
&(3, 100)    &(6, 582)    &0.8025  &0.7921                    &0.7819  &0.8051  &0.8139\\
&(3, 200)    &(6, 878)    &0.8022  &0.7905                    &0.7819  &\textbf{0.7963}  &\textbf{0.8051}\\
&(5, 100)    &(10, 379)   &0.7893  &0.7822                    &0.7817  &0.8013  &0.8131\\
&(5, 200)    &(10, 539)   &0.8036  &0.7978                    &0.7786  &0.7966  &0.8103\\
&(7, 100)    &(14, 320)   &0.7880  &0.7841                    &0.7822  &0.8046  &0.8090\\
&(7, 200)    &(14, 395)   &0.7950  &0.7875                    &\textbf{0.7775}  &0.8079  &0.8065\\
&average     &-           &0.7923  &0.7829                    &0.7774  &0.7960  &0.8026\\ \hline
\end{tabular}
\end{table*}

\begin{table*}[tbh]
\centering
\caption{STD performance of systems submitted by Participants of QUESST 2015}
\label{tab:cnxe_team}
\begin{tabular}{|l|l|l|l|}
\hline
Index & Methods    & actCnxe  & minCnxe \\ \hline
(1) & Caranica et al. Romanian Phones MFCC \cite{caranica2015speed}  & 1.0061 & 0.9944  \\ \hline
(2) & Caranica et al. Romanian Phones PNCC \cite{caranica2015speed}  & 1.0061 & 0.9943  \\ \hline
(3) & Ma et al. SMO+iSAX\cite{ma2015cuny}         & 0.9988 & 0.9872  \\ \hline
(4) & Ma et al. subseq+MFCC \cite{ma2015cuny}         & 1.0658 & 0.9823  \\ \hline
(5) & Sk{\'a}cel et al. Posteriorgrams DTW\cite{skacel2015but}      & 0.8452 & 0.8263  \\ \hline
(6) & Sk{\'a}cel et al. Posteriorgrams subsequence DTW\cite{skacel2015but}      & 0.8447 & 0.8124  \\ \hline
(7) & Hou et al.  Spectral, phoneme-state posterior, BNF, fusion of 66 systems \cite{hou2015nni}    & 0.773 & 0.757    \\ \hline
(8) & Proposed, Multi-granular Document Tokens Eq. \eqref{eq:dtw_p} (7,200)  & 0.9997 & 0.9937  \\ \hline
(9) & Proposed, Multi-granular Query Tokens Eq. \eqref{eq:dtw_p} Average & 1.0022 & 0.9965  \\ \hline
(10) & Proposed, Hierarchical Query Tokens Eq. \eqref{eq:dtw_p} Average & 1.0020 & 0.9964  \\ \hline
(11) & Proposed, Multi-granular Document Tokens Eq. \eqref{eq:dtw_p} Average & 1.0015 & 0.9932  \\ \hline
(12) & Proposed, Hierarchical Document Tokens Eq. \eqref{eq:dtw_p} Average & 1.0013 & 0.9932  \\ \hline
\end{tabular}
\end{table*}

\section{Spoken Term Detection Experiments}\label{sec:std}
We use the dataset provided by the “Query by Example Search on Speech Task” (QUESST), held as part of the MediaEval 2015 evaluation task \cite{szoke2015query}, in our spoken term detection experiments.
QUESST 2015 intended to evaluate language-independent audio search systems in a low resource scenario. 
The QUESST 2015 dataset is composed of a set of spoken documents, and 2 sets of spoken queries.
The spoken document set is composed of around 18 hours of audio (11662 files) in the following 7 languages: Albanian, Czech, English, Mandarin, Portuguese, Romanian and Slovak, with different amounts of audio per language.
The spoken queries, which are relatively short (5.8 seconds long on average), were automatically extracted from longer recordings and manually checked to avoid very short or very long utterances. 
The QUESST 2015 dataset includes 445 development queries and 447 evaluation queries, with the number of queries per language being more or less balanced with respect to the amount of audio available in the spoken document set.
Both of the two query sets contain three types of queries: the first one (T1) involves “exact matches” whereas the second one (T2) allows for inflectional variations of words or word re-orderings (that is, “approximate matches”); the third one (T3) is similar to T2, but the queries were drawn from conversational speech, thus containing strong coarticulations and some filler content between words.
The data was artificially noised and reverberated with equal amounts of clean, noisy, reverberated and noisy+reverberated speech. 
Reverberation was obtained by passing the audio through a filter with an artificially generated room impulse response (RIR).
The normalized cross entropy cost (Cnxe) \cite{rodriguez2013mediaeval,anguera2015quesst2014}, the lower the better, was used as the primary metric for the evaluation.

Below we only report the results on the development queries since the results were similar on the evaluation queries.
We trained four sets of token sets under the two proposed paradigms with the two testing scenarios (Document Tokens and Query Tokens), with different granularities. 
We list the model-dependent minCnxe \cite{szoke2015query}, obtained with either the Initial Relevance Score (IRS) in Eq. \eqref{eq:dtw} or Enhanced Relevance Score (ERS) in Eq. \eqref{eq:dtw_p} on the different token sets for Document Tokens and Query Tokens in Table \ref{tab:std_doc} and Table \ref{tab:std_qer} respectively. 
We further show the detailed results for ERS on different query types T1, T2, T3 respectively.
%
Results of other metrics are also available but they all showed consistent trends, therefore left out.
The model-dependent minCxne was selected because it showed the most variance across the different token sets.
In the top half of Tables \ref{tab:std_doc} and \ref{tab:std_qer}, we trained several Multi-granular token sets on the utterances with the granularities $(m,n)=(3,100),(3,200),(5,100),(5,200),(7,100),(7,200)$. 
%
%
%
%
In the bottom half of Table \ref{tab:std_doc}, the Hierarchical tokens are trained with lexicon and language models using the method in \cite{chung2013unsupervised}. 
In the bottom half of Table \ref{tab:std_qer}, we only used the lexicon and not the language model in the Hierarchical Paradigm because the noisy acoustic conditions did not allow the initialization step to find stable word-like structures and the queries are too short for an effective language models.
%
The number of states of the longest word-like token in the lexicon $u$, and the number of word-like tokens in the lexicon $v$ are also shown.
In training these token sets, we constrained the word-like tokens to be at most two subword-like tokens long, or $u\leq 2m$.
The model-dependent minCnxe results were evaluated using IRS in Eq. \eqref{eq:dtw} and ERS in Eq. \eqref{eq:dtw_p} based on the subword-like tokens with granularity $(m,n)$, except the decoding process for generating the token sequence representations was based on the lexicon constraints for the word-like tokens. 
%
%
%
%

In the last rows of the top and bottom halves of Tables \ref{tab:std_doc} and \ref{tab:std_qer} we averaged the relevance scores obtained at each granularity and evaluated the performance on the averaged relevance scores.
From Tables \ref{tab:std_doc} and \ref{tab:std_qer}, several observations can be made:
(a) In all cases, the results obtained with the Enhanced Relevance Scores (ERS) in Eq. \eqref{eq:dtw_p} performed better than results obtained with the Initial Relevance Score (IRS) in Eq. \eqref{eq:dtw}. 
These results verify  our analysis in Section \ref{sec:prince} regarding the heuristics behind token sequence representations.
This is a major improvement over our previous work which only considered IRS in Eq. \eqref{eq:dtw}.
(b) By comparing Tables \ref{tab:std_qer} and \ref{tab:std_doc}, the performance of some of the query tokens were comparable to the document tokens.
The query tokens at granularity (5,100) in Table \ref{tab:std_qer} even performed better than two document tokens at granularity (3,100), (3,200) in Table \ref{tab:std_doc} for ERS.
Only 445 short queries were used to train the query tokens, while 11662 long spoken documents were used to train the document tokens.
With only 445 short queries we trained $n=100$ or  $n=200$ token HMMs, so each HMM is given only a few training examples.
The comparable performance to the document tokens which were trained on 11662 long spoken documents suggests that under noisy conditions, our analysis in Section \ref{sec:use_qer} is probably correct.
Training HMM tokens on very small training sets is essentially just assigning the query features to the means of the Gaussians, and decoding the HMM on the documents is really just performing DTW on the query-document pairs.
(c) Eq. \eqref{eq:card_equib} successfully explains the trends between $m$, $n$, $u$ and $v$.
We constrained the word-like tokens in the lexicon to be at most two subword-like tokens long, so $u=2m$.
When we substitute $(m',n')$ with $(u,v)$, we have 
\begin{equation}
v=n^{\frac{u}{m}}, \label{eq:mnuv}
\end{equation}
which is the condition that the number of representations in terms of token sequences is saturated, and the two-level representation of the Hierarchical Paradigm is reduced to the one-level Multi-granular Paradigm.
Although the actual values of $v$ are way smaller than the theoretical value in Eq. \eqref{eq:mnuv} for saturation to happen, it actually explains some trends.
For the query tokens in Table \ref{tab:std_qer}, by comparing granularities (3,100), (5,100), (7,100) and (3,200), (5,200), (7,200) with a fixed $n=$100 or 200, we see that the actual value of $v$ decreases with the increase of $m$, which is consistent with Eq. \eqref{eq:mnuv}.
In other words, when the number of distinct subword-like tokens is fixed, longer subword-like tokens implies smaller number of distinct word-like tokens (smaller lexicon size).
This is only partially observed for the document tokens in Table \ref{tab:std_doc} when the growth of $v$ slows down.
By comparing the granularities (3,100) and (3,200),  (5,100) and (5,200), (7,100) and (7,200) with a fixed $m=$3, 5, 7 and we see that the actual value of $v$ increases with the increase of $n$, which is also consistent with Eq. \eqref{eq:mnuv}.
In other words, when the length of subword-like tokens is fixed more distinct subword-like token implies more distinct word-like tokens (larger lexicon size).
(d) Under most conditions, the Multi-granular Paradigm performed better than the Hierarchical Paradigm for both query tokens and document tokens.
We believe this is because careful tuning is required when training lexicons for the Hierarchical Paradigm to be successful.
(e) By comparing the performance of the averaged scores for different query types in both Tables \ref{tab:std_doc} and \ref{tab:std_qer}, T1$<$T2$<$T3. 
This indicating that T1 is the easiest type of query where T3 is the hardest.
(f) For both document tokens and query tokens, the Hierarchical tokens managed to get the best results on most individual query types T1, T2 and T3, but not the best when all 3 query types are considered.
This is probably because the Hierarchical tokens managed to capture some structure of the specific query types.
Good thresholds for scores can be derived for specific query types, but maybe the range of scores is too different across query types, degrading the performance when jointly considered.
%
%
%

For comparison to supervised methods, we also list the results of the systems submitted by participants of the QUESST 2015 evaluation on the evaluation set in Table \ref{tab:cnxe_team}.
Because most teams reported their results on the model-independent actual Cnxe (actCnxe) and minimum Cnxe (minCnxe), we also report our results in model-independent actCnxe and minCnxe in Table \ref{tab:cnxe_team} instead of the model-dependent minimum Cnxe in Tables \ref{tab:std_doc} and \ref{tab:std_qer}.
We list the results of our best performing token set in row (8) of Table \ref{tab:cnxe_team}, which is (7,200) of the Multi-granular document tokens in Table \ref{tab:std_doc}.
The results of the averaged relevance scores in Table \ref{tab:std_doc} and \ref{tab:std_qer} are also listed in rows (9), (10), (11), (12) in Table \ref{tab:cnxe_team}.
Note that in the QUESST 2015 evaluation the participants were allowed to use acoustic models trained on other labeled datasets, since the task did not require systems to compete under the zero resource scenario.
Also note that in our experiments we always assume that either the document or query is not available during the training of the token sets, so the comparison is not entirely fair.
In rows (1) and (2) Caranica et al. \cite{caranica2015speed} used supervised phoneme HMM tokens trained on a labeled Romanian corpus of 8.7 hours.
The results of row (1) were trained with MFCCs, and row (2) with PNCCs \cite{kelly2010comparison}.
Their systems are similar to the proposed approach here because we both use token HMMs.
Although they used supervised knowledge in their HMMs, the performance of our unsupervised HMMs in row (11) and (12) is actually better.
In rows (3) and (4) Ma et al. \cite{ma2015cuny} used various combinations of Czech, Hungarian, and Russian phonetic tokens and frame-based DTW systems.
Their systems are similar to the proposed approaches here because we both used the fusion of various token sets.
In row (3), they fused the results from various ways to the calculate distances based on the phone sequences for three different languages.
In row (4), they combined the distances of the phonetic tokens with frame-based DTW.
Their supervised results are comparable to ours.
In rows (5) and (6) Sk{\'a}cel et al. performed frame-based DTW on posteriorgrams extracted from Czech, Portuguese, Russian, Spanish systems. 
In row (5), they stacked the posteriorgrams and performed DTW.
In row (6), they split the queries into multiple segments and performed DTW on each segment then averaged the results.
The rationale behind this approach of splitting the queries into segments in row (6) is because of the complications of the T2 and T3 queries.
The improvement from rows (5) to (6) suggests that this action is justified.
We did not develop any special approach to deal with the T2 and T3 queries, which may explain their better results.
In row (7), Hou et al. fused 66 systems of spectral features, phoneme-state posterior features and bottleneck features from 3 teams.
The performance of their aggregated system  was the best performing by a large margin and can be considered the topline of QUESST 2015.

Considering real applications, the major advantage of unsupervised approaches as proposed here over the multilingual training (using supervised models trained on other languages) is the robustness across languages. The performance of multilingual approaches have been observed in experiments to rely on how closely the linguistic structure for the given corpora of the language is related to that of the languages for the supervised models. In addition, the given corpora may be multilingual with code-switching, which makes robustness across languages even more important. This is also important for models endangered languages like the 26 Formosan languages\cite{fox2004current,zeitoun2005formosan}. For such endangered languages, the supervised models from which we can borrow related linguistic structures may not exist.

\section{Subword-like Tokens Evaluation}\label{sec:abx}
We use the corpus and evaluation defined in the Zero Resource Speech Challenge 2015 \cite{versteegh2015zero} to evaluate quality of the sub-word like tokens obtained in this work.
We choose Track 1 of the Challenge to evaluate the quality of the sub-word structures on two languages, English and Xitsonga.
%
%
The Track 1 evaluation was based on the ABX discriminability test \cite{schatz2013evaluating} including across-speaker and within-speaker tests. 
The warping distance obtained by performing DTW over the sequences of the obtained frame-level features for predefined signal pairs were used as the distance metric for the ABX discriminability test.
For the test here, we use posteriorgrams with dimensionality $n$ ($n$ is the number of distinct subword-like tokens as used above) extracted from the decoded token lattices as the features to be evaluated in this experiment.
There is no separate query set, only document tokens were considered.
Because the durations of the predefined signal pairs for the test were short and designed to evaluate frame-level speech features, the subword-tokens extracted from the paradigms can be as long as or longer than the the entire duration of the signal.
Since the scenario is quite different from the original design of the challenge, the metrics of the challenge is used to compare the different granularities and the paradigms rather than taken as a quality measure.
The results in error percentage (the lower the better) on English and Xitsonga is listed in Table \ref{tab:abx_eng} and Table \ref{tab:abx_xit} respectively. Table \ref{tab:abx_baseline}, also lists the performance of other systems as references.

In Table \ref{tab:abx_eng} and Table \ref{tab:abx_xit}, we trained acoustic tokens using the Multi-granular Paradigm and the Hierarchical Paradigm on the two spoken archives in English and Xitsonga respectively.
The results on both languages have similar trends and some observations can be made.
(a) The shorter the HMM, the better the performance, probably because shorter HMMs can better fit the short-time variation of the signals for the evaluation intervals defined by the task.
(b) Most results under the Hierarchical Paradigm were in general better than those under the Multi-granular Paradigm, which is the opposite of observation (d) of Section \ref{sec:std}.
This is probably because unlike MediaEval QUESST 2015, the Zero Resource Challenge 2015 used clean speech and did not mix multiple languages.
With less noise, the Hierarchical Paradigm can better capture longer word-like tokens, leading to better subword-like tokens.
For MediaEval QUESST 2015, the noise and different language structures made it difficult to build word-like tokens from subword-like tokens in the Hierarchical Paradigm.

In Table \ref{tab:abx_baseline}, we compare the best system at granularity (3,50) for both the Multi-granular Paradigm and Hierarchical Paradigm with the performance of other systems reported by the Challenge.
The official baseline provided by the Challenge was the MFCC features without delta and double delta and the official topline was supervised phone posteriorgrams. 
The system proposed by Thiolli\`ere et al. \cite{thiolliere2015hybrid} had two components: a dynamic-time warping (DTW) based spoken term discovery (STD) system and a Siamese DNN. 
The STD system clustered word-sized repeated fragments in the acoustic streams while the DNN was trained to minimize the distance between time aligned frames of tokens of the same cluster, and maximize the distance between tokens of different clusters.
The frame-level features were then extracted from the bottleneck layer of the trained DNN.
Renshaw et al. \cite{renshaw2015comparison} proposed a similar system using correspondence autoencoders (cAE). 
The cAE was an autoencoder trained on feature pairs, one feature as the input and the other as the reconstruction target at the output. 
The frame-level features were then extracted from the bottleneck layer of the trained cAE.
Like the hybrid Siamese system above \cite{thiolliere2015hybrid}, a DTW based system was used to align the feature pairs for feature sequences within the same cluster.
The clusters can either be ground truth word types or discovered clusters.
The performance was better when the clusters were the ground truth word types, although in that case the cAE/hybrid Siamese system was not an unsupervised model.
Badino et al. \cite{badino2015discovering} proposed the generation of discrete features by forcing the bottleneck features of autoencoders (AE) to be binary.
The binary bottleneck features with dimensionality $H$ extracted from the AE could be interpreted as an integer between $0$ and $2^H-1$. 
The discrete integer sequence was further refined with HMMs.
%
%
Chen et al. \cite{chen2015parallel} used a Dirichlet process Gaussian mixture model (DPGMM) to represent speech frames with Gaussian posteriorgrams. 
The model performed unsupervised clustering on untranscribed data, and each Gaussian component could be considered as a cluster of sounds from various speakers. 
The model inferred its model complexity (i.e. the number of Gaussian components) directly from the data.
%
%
%
Baljekar et al. \cite{baljekar2015using} used Articulatory Features (AF) trained on labeled speech in a higher resource language to infer phonological segments of varying granularity.
Both the frame-level AFs and the token-like inferred phonological units were used in the evaluation.
The results of our system are listed for reference. 

\begin{table}[t]
\centering
\caption{ABX performance of Multi-granular Tokens and Hierarchical Tokens at different granularities trained on the English Corpus of the Zero Resource Speech Challenge 2015}
\label{tab:abx_eng}
\begin{tabular}{|c|c|c|c|c|}
\hline
Paradigm& (m, n) & (u, v) & within & across \\ \hline\multirow{6}{*}{Multi-granular}
&(3, 50)     &-           &25.52  &16.73\\
&(3, 100)    &-           &27.64  &17.86\\
&(5, 50)     &-           &26.36  &17.14\\
&(5, 100)    &-           &27.98  &17.78\\
&(7, 50)     &-           &27.47  &18.25\\
&(7, 100)    &-           &29.32  &18.82\\ \hline
\multirow{6}{*}{Hierarchical}
&(3, 50)     &(6, 698)    &\textbf{25.29}  &\textbf{16.50}\\
&(3, 100)    &(6, 2036)   &27.40  &17.63\\
&(5, 50)     &(10, 1001)  &26.36  &17.14\\
&(5, 100)    &(10, 2336)  &27.98  &17.78\\
&(7, 50)     &(14, 1176)  &27.41  &18.18\\
&(7, 100)    &(14, 2248)  &29.32  &18.82\\ \hline
\end{tabular}
\end{table}

\begin{table}[t]
\centering
\caption{ABX performance of Multi-granular Tokens and Hierarchical Tokens at different granularities trained on the Xitsonga Corpus of the Zero Resource Speech Challenge 2015}
\label{tab:abx_xit}
\begin{tabular}{|c|c|c|c|c|}
\hline
Paradigm & (m, n) & (u, v) & within & across \\ \hline
\multirow{6}{*}{Multi-granular}
&(3, 50)     &-           &\textbf{23.92}  &14.75\\
&(3, 100)    &-           &25.77  &15.34\\
&(5, 50)     &-           &25.16  &16.20\\
&(5, 100)    &-           &27.38  &16.65\\
&(7, 50)     &-           &26.52  &17.45\\
&(7, 100)    &-           &27.89  &17.93\\ \hline
\multirow{6}{*}{Hierarchical}
&(3, 50)     &(6, 549)    &23.98  &\textbf{14.67}\\
&(3, 100)    &(6, 1185)   &25.88  &15.24\\
&(5, 50)     &(10, 701)   &25.14  &16.31\\
&(5, 100)    &(10, 1213)  &27.11  &16.81\\
&(7, 50)     &(14, 705)   &26.51  &17.47\\
&(7, 100)    &(14, 1012)  &27.82  &17.87\\ \hline
\end{tabular}
\end{table}
\begin{table}[tbh]
\centering
\caption{ABX performance of systems submitted by participants of the Zero Resource Speech Challenge 2015}
\label{tab:abx_baseline}
\begin{tabular}{|l|r|r|r|r|}
\hline
\multicolumn{1}{|l|}{\multirow{2}{*}{Method}} & \multicolumn{2}{c|}{English}                              & \multicolumn{2}{c|}{Xitsonga}                             \\ \cline{2-5} 
\multicolumn{1}{|c|}{}                        & \multicolumn{1}{c|}{across} & \multicolumn{1}{l|}{within} & \multicolumn{1}{c|}{across} & \multicolumn{1}{c|}{within} \\ \hline
Topline           								& 16.0    & 12.1    & 4.5     & 3.5          \\ \hline
Baseline               							& 28.1    & 15.6    & 33.8     & 19.1          \\ \hline
Thiolli\`ere et al. \cite{thiolliere2015hybrid} & 17.9    & 12.0    & 16.6      & 11.7         \\ \hline
Renshaw et al. \cite{renshaw2015comparison}     & 21.1    & 13.5    & 19.3      & 11.9         \\ \hline
Badino et al. \cite{badino2015discovering}      & 26.3    & 17.3    & 23.6      & 14.1         \\ \hline
Chen et al. \cite{chen2015parallel}    			& 16.3   & 10.8    & 17.2      & 9.6         \\ \hline
Baljekar et al. \cite{baljekar2015using}     	& 29.8    & 18.4    & 29.7      & 18.1         \\ \hline
Proposed (3,50) Mult.      						& 25.5     & 16.7    & 24.0      & 14.8         \\ \hline
Proposed (3,50) Hier.     						& 25.3    & 16.5    & 24.0       & 14.7        \\ \hline
\end{tabular}
\end{table}

\section{Choosing between the two Paradigms}\label{sec:support}
The goal of the experiments above is to provide a side-by-side comparison of the two paradigms on the same tasks, but the true strength of the idea of having two paradigms lies in choosing which to use for a given task. 
For a given task, usually one paradigm would be preferred over the other and they would seldom be used together.
The experimental results show that the Hierarchical Paradigm can achieve the best performance at the correct granularities, since linguistic structures provide context for the acoustic tokens. 
However, finding the correct granularities usually involves a grid search over the hyperparameter space which could be done more easily by training acoustic tokens under the Multi-granular Paradigm.
With the acoustic subword-like tokens alone, the Multi-granular Paradigm can achieve decent performance by simply aggregating the scores of multiple token sets.

If a task has constraints on computation power so decoding with a large lexicon under the Hierarchical Paradigm becomes difficult, or if the task has to be robust across various acoustic conditions, we can simply take the average scores from the multiple token sets of the Multi-granular Paradigm and ignore the hierarchical structures.
For example, we have shown that supervised speaker-independent DNNs adapted with unsupervised speaker-dependent Multi-granular tokens as auxiliary targets can be used for speaker adaptation \cite{wei2017personalized}.
By training Multi-granular tokens directly on the audio of a specific user, the system could capture speaker-specific acoustic tokens that could be due to dialect.
The high degree of similarity between the HMMs of the multiple sets of unsupervised tokens under the Multi-granular Paradigm and the supervised speaker independent phoneme models make it possible for them to learn from each other through the shared layers of the DNN.
%
%
The proposed semi-supervised approach has beaten strong adaptation baselines.

If a task has constraints on storage space making it difficult to store multiple representations using the Multi-granular Paradigm, or if the task requires only one high quality representation for every audio file, we can select a few promising granularities to train the Hierarchical Paradigm and discard the rest.
For example, we have shown that the quality of the hierarchical tokens can be good enough for query expansion in semantic retrieval of spoken content \cite{li2013towards,lee2013enhancing}.  
%
%
A text-based query expansion retrieval system returns documents containing exact matches in the first pass of the system for a given query.
Words that appear frequently in the retrieved documents are considered to be semantically related, and treated as expanded queries.
In the second pass, the system also returns documents containing expanded queries.
Using the Hierarchical Paradigm, this technique can be applied to spoken documents as well by treating the acoustic tokens as regular words.
Many Out-of-Vocabulary (OOV) words incorrectly recognized by ASR systems can be consistently represented by acoustic tokens.
For an unannotated spoken corpus, the user can say ``President", and the system would return spoken documents containing ``Roosevelt" without any knowledge of the content.

Table \ref{tab:summary} is a summary of the computation complexity for the Multi-granular Paradigm and Hierarchical Paradigm based on the notations explained in Table \ref{tab:notation}.
Only the decoding step in Eq. \eqref{eq:3} is different for the two Paradigms.
The explanations of the content of Table \ref{tab:summary} is in Appendix \ref{sec:complexity}. 
%
%
These Tables can be used as a reference for estimating the resources required and deciding which of the two paradigms to use.

\section{Conclusion} \label{sec:conclusion}
This paper presents two different paradigms for unsupervised discovery of structured acoustic tokens from a given spoken corpus, in which the acoustic tokens discovered are structured in two different ways.
In the Multi-granular Paradigm, we are able to discover many sets of acoustic tokens over a two-dimensional space of temporal granularity and phonetic granularity, and these set of tokens can be complementary to each other.
In the Hierarchical Paradigm, the two-level word-like and subword-like tokens are learned layer after layer with the proposed cascaded stages of iterative optimization.
We then unify the two paradigms in a single theoretical framework, and discuss when it would be better to choose one over the other.
We performed Spoken Term Detection experiments on the MediaEval QUESST 2015 corpus and ABX evaluation on the Zero Resource Challenge 2015 corpus to verify the competitiveness of the discovered acoustic tokens.

\begin{table*}[tbh]
\centering
\caption{Summary of computation complexity for Multi-granular tokens with granularity parameters $(m,n)$ and Hierarchical tokens with lexical parameters $(u,v)$ using the notations in Table \ref{tab:notation}.}
\label{tab:summary}
\begin{tabular}{|l|l|l|l|} \hline
Operation                 & Equation/Notation &  Time				& Storage			\\ \hline
Token Training            &Eq. \eqref{eq:2}& $O(UT^2gf/d)$       &   			          	\\ 
Token Acoustic Model      &$\theta^{mn}$            &           				& $O(mngf)$             	\\
Token Lexical Model       &$\theta_{uv}$            &           				& $O(v\log v +uv \log n)$             	\\ 
Multi-granular Token Decoding       &Eq. \eqref{eq:3} & $O(UTmn^2gf)$         		&          					\\ 
Hierarchical Token Decoding   &Eq. \eqref{eq:3} & $O(UTmn^{2u/m}gf)$         		&          					\\ 
Token Sequence      &$W$            &           				& $O(UT\log(n)/md)$        	\\ 
Token Distance      &Eq. \eqref{eq:soft}& $O(mn^2g^2f)$           	& $O(n^2)$             	\\ 
Token DTW           &DTW for $W$& $O(U_sU_qT_sT_q/m^2d^2)$        &              				\\ \hline
Feature Sequence    & $X$ &           				& $O(UTf)$             	\\ 
Feature DTW         &DTW for $X$& $O(U_sU_qT_sT_qf)$         		&              				\\ \hline
Storage Compression         &$X/(\theta + W)$&         		&  $O(fmd/\log n)$     \\ 		
Time Compression         &DTW for $X$/ DTW for $W$ & $O(fm^2d^2)$          		&     		\\ \hline
\end{tabular}
\end{table*}

\begin{table*}[tbh]
\centering
\caption{Summary of notations used in Table \ref{tab:summary}.}
\label{tab:notation}
\begin{tabular}{|l|l|} \hline
Notation & Definition								\\ \hline
$W$		&decoded token sequences at the $i$th iteration\\ 
$X$		&the frame-level acoustic feature sequence\\ 
$\theta$	&general parameters of the model\\ 
$\theta^{mn}$	&the HMM parameters of $n$ HMMs, each being $m$ states\\ 
$\theta_{uv}$	&the lexicon containing $v$ sequences of HMMs, with the longest being $u$ states long \\  
$\theta^{mn}_{uv}$	&the parameter set $\{\theta^{mn}, \theta_{uv}\}$\\ \hline
$m$		&the number of states      					\\ 
$n$		&the number of HMMs      					\\ 
$u$		&the number of states in the longest word  in the lexicon		\\ 
$v$		&the number of words in the lexicon   					\\ 
$g$		&the number of Gaussians in each state		\\ 
$f$		&the rank of the feature					\\
$d$		&the average duration of each state 		\\
$T$		&the average duration of each utterance 	\\
$T_s$		&the average duration of each utterance in the document corpus 	\\
$T_q$		&the average duration of each utterance in the query corpus	\\
$U$		    &the number of utterances 				\\
$U_s$		&the number of utterances in the document corpus 				\\
$U_q$		&the number of utterances in the query corpus 				\\
\hline
\end{tabular}
\end{table*}

\bibliographystyle{IEEEbib}
\bibliography{mycap}

\begin{thebibliography}{10}

\bibitem{gales2014speech}
Mark~JF Gales, Kate~M Knill, Anton Ragni, and Shakti~P Rath,
\newblock ``Speech recognition and keyword spotting for low-resource languages:
  Babel project research at cued.,''
\newblock in {\em SLTU}, 2014, pp. 16--23.

\bibitem{chan2011unsupervised}
Chun-An Chan and Lin-Shan Lee,
\newblock ``Unsupervised hidden markov modeling of spoken queries for spoken
  term detection without speech recognition.,''
\newblock in {\em INTERSPEECH}, 2011, pp. 2141--2144.

\bibitem{huijbregts2011unsupervised}
Marijn Huijbregts, Mitchell McLaren, and David van Leeuwen,
\newblock ``Unsupervised acoustic sub-word unit detection for query-by-example
  spoken term detection,''
\newblock in {\em Acoustics, Speech And Signal Processing (ICASSP), {2}011 IEEE
  International Conference On}. IEEE, 2011, pp. 4436--4439.

\bibitem{siu2014unsupervised}
Man-hung Siu, Herbert Gish, Arthur Chan, William Belfield, and Steve Lowe,
\newblock ``Unsupervised training of an {HMM}-based self-organizing unit
  recognizer with applications to topic classification and keyword discovery,''
\newblock {\em Computer Speech \& Language}, vol. 28, no. 1, pp. 210--223,
  2014.

\bibitem{lee2012nonparametric}
Chia-ying Lee and James Glass,
\newblock ``A nonparametric bayesian approach to acoustic model discovery,''
\newblock in {\em Proceedings Of The {5}0th Annual Meeting Of The Association
  For Computational Linguistics: Long Papers-Volume {1}}. Association for
  Computational Linguistics, 2012, pp. 40--49.

\bibitem{novotney2009unsupervised}
Scott Novotney, Richard Schwartz, and Jeff Ma,
\newblock ``Unsupervised acoustic and language model training with small
  amounts of labelled data,''
\newblock in {\em Acoustics, Speech and Signal Processing, 2009. ICASSP 2009.
  IEEE International Conference on}. IEEE, 2009, pp. 4297--4300.

\bibitem{wei2017personalized}
Cheng-Kuan Wei, Cheng-Tao Chung, Hung-Yi Lee, and Lin-Shan Lee,
\newblock ``Personalized acoustic modeling by weakly supervised multi-task deep
  learning using acoustic tokens discovered from unlabeled data,''
\newblock submitted for a future conference.

\bibitem{kamper2017segmental}
Herman Kamper, Aren Jansen, and Sharon Goldwater,
\newblock ``A segmental framework for fully-unsupervised large-vocabulary
  speech recognition,''
\newblock {\em Computer Speech \& Language}, 2017.

\bibitem{szoke2015copingwith}
Igor Sz{\"o}ke, Miroslav Sk{\'a}cel, Luk{\'a}{\v{s}} Burget, and Jan
  {\v{C}}ernock{\`y},
\newblock ``Coping with channel mismatch in query-by-example-but quesst 2014,''
\newblock in {\em Acoustics, Speech and Signal Processing (ICASSP), 2015 IEEE
  International Conference on}. IEEE, 2015, pp. 5838--5842.

\bibitem{leung2016toward}
Cheung-Chi Leung, Lei Wang, Haihua Xu, Jingyong Hou, Van~Tung Pham, Hang Lv,
  Lei Xie, Xiong Xiao, Chongjia Ni, Bin Ma, et~al.,
\newblock ``Toward high-performance language-independent query-by-example
  spoken term detection for mediaeval 2015: Post-evaluation analysis,''
\newblock in {\em Proc. INTERSPEECH}, 2016.

\bibitem{chen2016unsupervised}
Hongjie Chen, Cheung-Chi Leung, Lei Xie, Bin Ma, and Haizhou Li,
\newblock ``Unsupervised bottleneck features for low-resource query-by-example
  spoken term detection,''
\newblock in {\em Proc. INTERSPEECH}, 2016.

\bibitem{yang2014intrinsic}
Peng Yang, Cheung-Chi Leung, Lei Xie, Bin Ma, and Haizhou Li,
\newblock ``Intrinsic spectral analysis based on temporal context features for
  query-by-example spoken term detection.,''
\newblock in {\em INTERSPEECH}, 2014, pp. 1722--1726.

\bibitem{wang2015acoustic}
Haipeng Wang, Tan Lee, Cheung-Chi Leung, Bin Ma, and Haizhou Li,
\newblock ``Acoustic segment modeling with spectral clustering methods,''
\newblock {\em IEEE/ACM Transactions on Audio, Speech and Language Processing
  (TASLP)}, vol. 23, no. 2, pp. 264--277, 2015.

\bibitem{renshaw2015comparison}
Daniel Renshaw, Herman Kamper, Aren Jansen, and Sharon Goldwater,
\newblock ``A {C}omparison of {N}eural {N}etwork {M}ethods for {U}nsupervised
  {R}epresentation {L}earning on the {Z}ero {R}esource {S}peech {C}hallenge,''
\newblock in {\em Proceedings of Interspeech}, 2015.

\bibitem{zhang2013unsupervised}
Yaodong Zhang,
\newblock {\em Unsupervised {S}peech {P}rocessing with {A}pplications to
  query-by-example {S}poken {T}erm {D}etection},
\newblock Ph.D. thesis, Massachusetts Institute of Technology, 2013.

\bibitem{huang2015rapid}
Zhen Huang, Jinyu Li, Sabato~Marco Siniscalchi, I-Fan Chen, Ji~Wu, and Chin-Hui
  Lee,
\newblock ``Rapid adaptation for deep neural networks through multi-task
  learning.,''
\newblock in {\em Interspeech}, 2015, pp. 3625--3629.

\bibitem{park2008unsupervised}
Alex~S Park and James~R Glass,
\newblock ``Unsupervised pattern discovery in speech,''
\newblock {\em IEEE Transactions on Audio, Speech, and Language Processing},
  vol. 16, no. 1, pp. 186--197, 2008.

\bibitem{jansen2012indexing}
Aren Jansen and Benjamin Van~Durme,
\newblock ``Indexing raw acoustic features for scalable zero resource
  search.,''
\newblock in {\em INTERSPEECH}, 2012.

\bibitem{jansen2011towards}
Aren Jansen and Kenneth Church,
\newblock ``Towards unsupervised training of speaker independent acoustic
  models.,''
\newblock in {\em INTERSPEECH}, 2011, pp. 1693--1692.

\bibitem{gish2009unsupervised}
Herbert Gish, Man-hung Siu, Arthur Chan, and William Belfield,
\newblock ``Unsupervised training of an hmm-based speech recognizer for topic
  classification.,''
\newblock in {\em INTERSPEECH}, 2009, pp. 1935--1938.

\bibitem{chung2013unsupervised}
Cheng-Tao Chung, Chun-an Chan, and Lin-shan Lee,
\newblock ``Unsupervised discovery of linguistic structure including two-level
  acoustic patterns using three cascaded stages of iterative optimization,''
\newblock in {\em Acoustics, Speech And Signal Processing (ICASSP), {2}013 IEEE
  International Conference On}. IEEE, 2013, pp. 8081--8085.

\bibitem{chung2014unsupervised}
Cheng-Tao Chung, Chun-an Chan, and Lin-shan Lee,
\newblock ``Unsupervised spoken term detection with spoken queries by
  multi-level acoustic patterns with varying model granularity,''
\newblock in {\em Acoustics, Speech And Signal Processing (ICASSP), {2}014 IEEE
  International Conference On}. IEEE, 2014.

\bibitem{chung2015enhancing}
Cheng-Tao Chung, Wei-Ning Hsu, Cheng-Yi Lee, and Lin-Shan Lee,
\newblock ``Enhancing automatically discovered multi-level acoustic
  patternsconsidering context {C}onsistency with applications in spoken term
  detection,''
\newblock in {\em Acoustics, Speech And Signal Processing (ICASSP), {2}015 IEEE
  International Conference On}. IEEE, 2015.

\bibitem{li2013towards}
Yun-Chiao Li, Hung-yi Lee, Cheng-Tao Chung, Chun-an Chan, and Lin-shan Lee,
\newblock ``Towards unsupervised semantic retrieval of spoken content with
  query expansion based on automatically discovered acoustic patterns,''
\newblock in {\em Automatic Speech Recognition and Understanding (ASRU), 2013
  IEEE Workshop on}. IEEE, 2013, pp. 198--203.

\bibitem{zhang2010towards}
Yaodong Zhang and James~R Glass,
\newblock ``Towards multi-speaker unsupervised speech pattern discovery,''
\newblock in {\em 2010 IEEE International Conference on Acoustics, Speech and
  Signal Processing}. IEEE, 2010, pp. 4366--4369.

\bibitem{chen2015parallel}
Hongjie Chen, Cheung-Chi Leung, Lei Xie, Bin Ma, and Haizhou Li,
\newblock ``{P}arallel {I}nference of {D}irichlet {P}rocess {G}aussian
  {M}ixture {M}odels for {U}nsupervised {A}coustic {M}odeling: A feasibility
  study,''
\newblock in {\em Proceedings of Interspeech}, 2015.

\bibitem{kamperunsupervised}
Herman Kamper, Micha Elsner, Aren Jansen, and Sharon Goldwater,
\newblock ``Unsupervised neural network based feature extraction using weak
  top-down constraints,''
\newblock .

\bibitem{chung2015iterative}
Cheng-Tao Chung, Cheng-Yu Tsai, Hsiang-Hung Lu, Chia-Hsiang Liu, Hung-yi Lee,
  and Lin-shan Lee,
\newblock ``An iterative deep learning framework for unsupervised discovery of
  speech features and linguistic units with applications on spoken term
  detection,''
\newblock in {\em Automatic Speech Recognition and Understanding (ASRU), 2015
  IEEE Workshop on}. IEEE, 2015, pp. 245--251.

\bibitem{lee2013enhancing}
Hung-yi Lee, Yun-Chiao Li, Cheng-Tao Chung, and Lin-shan Lee,
\newblock ``Enhancing query expansion for semantic retrieval of spoken content
  with automatically discovered acoustic patterns,''
\newblock in {\em Acoustics, Speech And Signal Processing (ICASSP), {2}013 IEEE
  International Conference On}. IEEE, 2013, pp. 8297--8301.

\bibitem{miller2007rapid}
David~RH Miller, Michael Kleber, Chia-Lin Kao, Owen Kimball, Thomas Colthurst,
  Stephen~A Lowe, Richard~M Schwartz, and Herbert Gish,
\newblock ``Rapid and accurate spoken term detection.,''
\newblock in {\em INTERSPEECH}, 2007, pp. 314--317.

\bibitem{mamou2007vocabulary}
Jonathan Mamou, Bhuvana Ramabhadran, and Olivier Siohan,
\newblock ``Vocabulary independent spoken term detection,''
\newblock in {\em Proceedings Of The {3}0th Annual International ACM SIGIR
  Conference On Research And Development In Information Retrieval}. ACM, 2007,
  pp. 615--622.

\bibitem{wallace2007phonetic}
Roy~G Wallace, Robert~J Vogt, and Sridha Sridharan,
\newblock ``A phonetic search approach to the {2}006 nist spoken term detection
  evaluation,''
\newblock 2007.

\bibitem{pan2010performance}
Yi-Cheng Pan and Lin-shan Lee,
\newblock ``Performance analysis for lattice-based speech indexing approaches
  using words and subword units,''
\newblock {\em Audio, Speech, and Language Processing, IEEE Transactions on},
  vol. 18, no. 6, pp. 1562--1574, 2010.

\bibitem{saraclar2004lattice}
Murat Saraclar and Richard Sproat,
\newblock ``Lattice-based search for spoken utterance retrieval,''
\newblock {\em Urbana}, vol. 51, pp. 61801, 2004.

\bibitem{boves2009resources}
Lou Boves, Rolf Carlson, Erhard~W Hinrichs, David House, Steven Krauwer, Lothar
  Lemnitzer, Martti Vainio, and Peter Wittenburg,
\newblock ``Resources for speech research: Present and future infrastructure
  needs.,''
\newblock in {\em INTERSPEECH}. Citeseer, 2009, pp. 1803--1806.

\bibitem{kumar2007wwtw}
Arun Kumar, Nitendra Rajput, Dipanjan Chakraborty, Sheetal~K Agarwal, and
  Amit~A Nanavati,
\newblock ``Wwtw: The world wide telecom web,''
\newblock in {\em Proceedings Of The {2}007 Workshop On Networked Systems For
  Developing Regions}. ACM, 2007, p.~7.

\bibitem{carlin2011rapid}
Michael~A Carlin, Samuel Thomas, Aren Jansen, and Hynek Hermansky,
\newblock ``Rapid evaluation of speech representations for spoken term
  discovery.,''
\newblock in {\em INTERSPEECH}, 2011, pp. 821--824.

\bibitem{zhang2009unsupervised}
Yaodong Zhang and James~R Glass,
\newblock ``Unsupervised spoken keyword spotting via segmental dtw on gaussian
  posteriorgrams,''
\newblock in {\em Automatic Speech Recognition \& Understanding, {2}009. ASRU
  {2}009. IEEE Workshop On}. IEEE, 2009, pp. 398--403.

\bibitem{wang2012acoustic}
Haipeng Wang, Cheung-Chi Leung, Tan Lee, Bin Ma, and Haizhou Li,
\newblock ``An acoustic segment modeling approach to query-by-example spoken
  term detection,''
\newblock in {\em Acoustics, Speech And Signal Processing (ICASSP), {2}012 IEEE
  International Conference On}. IEEE, 2012, pp. 5157--5160.

\bibitem{zhang2011piecewise}
Yaodong Zhang and James~R Glass,
\newblock ``A piecewise aggregate approximation lower-bound estimate for
  posteriorgram-based dynamic time warping.,''
\newblock in {\em INTERSPEECH}, 2011, pp. 1909--1912.

\bibitem{zhang2012fast}
Yaodong Zhang, Kiarash Adl, and James Glass,
\newblock ``Fast spoken query detection using lower-bound dynamic time warping
  on graphical processing units,''
\newblock in {\em Acoustics, Speech And Signal Processing (ICASSP), {2}012 IEEE
  International Conference On}. IEEE, 2012, pp. 5173--5176.

\bibitem{szoke2015query}
Igor Sz{\"o}ke, Luis~Javier Rodr{\'\i}guez-Fuentes, Andi Buzo, Xavier Anguera,
  Florian Metze, Jorge Proenca, Martin Lojka, and Xiao Xiong,
\newblock ``Query by example search on speech at mediaeval 2015.,''
\newblock in {\em MediaEval}, 2015.

\bibitem{versteegh2015zero}
Maarten Versteegh, Roland Thiolliere, Thomas Schatz, Xuan~Nga Cao, Xavier
  Anguera, Aren Jansen, and Emmanuel Dupoux,
\newblock ``The zero resource speech challenge 2015,''
\newblock in {\em Proc. of INTERSPEECH}, 2015.

\bibitem{ong1999updateable}
Thian-Huat Ong and Hsinchun Chen,
\newblock ``Updateable {PAT-Tree} approach to chinese key phrase extraction
  using mutual information: A linguistic foundation for knowledge management,''
\newblock 1999.

\bibitem{hershey2007approximating}
John~R Hershey and Peder~A Olsen,
\newblock ``Approximating the kullback leibler divergence between gaussian
  mixture models,''
\newblock in {\em Acoustics, Speech And Signal Processing, {2}007. ICASSP
  {2}007. IEEE International Conference On}. IEEE, 2007, vol.~4, pp. IV--317.

\bibitem{juang1984hidden}
B-H Juang,
\newblock ``On the hidden markov model and dynamic time warping for speech
  recognition—a unified view,''
\newblock {\em Bell Labs Technical Journal}, vol. 63, no. 7, pp. 1213--1243,
  1984.

\bibitem{caranica2015speed}
Alexandru Caranica, Andi Buzo, Horia Cucu, and Corneliu Burileanu,
\newblock ``Speed@ mediaeval 2015: Multilingual phone recognition approach to
  query by example std.,''
\newblock in {\em MediaEval}, 2015.

\bibitem{ma2015cuny}
Min Ma and Andrew Rosenberg,
\newblock ``Cuny systems for the query-by-example search on speech task at
  mediaeval 2015.,''
\newblock in {\em MediaEval}, 2015.

\bibitem{skacel2015but}
Miroslav Sk{\'a}cel and Igor Sz{\"o}ke,
\newblock ``But quesst 2015 system description.,''
\newblock in {\em MediaEval}, 2015.

\bibitem{hou2015nni}
Jingyong Hou, Cheung-Chi~Leung Van Tung~Pham, Cheung-Chi Leung, Lei Wang,
  Haihua Xu, Hang Lv, Lei Xie, Zhonghua Fu, Chongjia Ni, Xiong Xiao, et~al.,
\newblock ``The nni query-by-example system for mediaeval 2015.,''
\newblock in {\em MediaEval}, 2015.

\bibitem{rodriguez2013mediaeval}
Luis~J Rodriguez-Fuentes and Mikel Penagarikano,
\newblock ``Mediaeval 2013 spoken web search task: system performance
  measures,''
\newblock 2013.

\bibitem{anguera2015quesst2014}
Xavier Anguera, Luis-J Rodriguez-Fuentes, Andi Buzo, Florian Metze, Igor
  Sz{\"o}ke, and Mikel Penagarikano,
\newblock ``Quesst2014: evaluating query-by-example speech search in a
  zero-resource setting with real-life queries,''
\newblock in {\em Acoustics, Speech and Signal Processing (ICASSP), 2015 IEEE
  International Conference on}. IEEE, 2015, pp. 5833--5837.

\bibitem{kelly2010comparison}
Finnian Kelly and Naomi Harte,
\newblock ``A comparison of auditory features for robust speech recognition,''
\newblock in {\em Signal Processing Conference, 2010 18th European}. IEEE,
  2010, pp. 1968--1972.

\bibitem{fox2004current}
James Fox et~al.,
\newblock ``Current developments in comparative austronesian studies,''
\newblock 2004.

\bibitem{zeitoun2005formosan}
Elizabeth Zeitoun and Ching-Hua Yu,
\newblock ``The formosan language archive: Linguistic analysis and language
  processing,''
\newblock {\em Computational Linguistics and Chinese Language Processing}, vol.
  10, no. 2, pp. 167--200, 2005.

\bibitem{schatz2013evaluating}
Thomas Schatz, Vijayaditya Peddinti, Francis Bach, Aren Jansen, Hynek
  Hermansky, and Emmanuel Dupoux,
\newblock ``Evaluating speech features with the minimal-pair {ABX} task:
  Analysis of the classical {MFC/PLP} pipeline,''
\newblock in {\em INTERSPEECH {2}013: {1}4th Annual Conference Of The
  International Speech Communication Association}, 2013, pp. 1--5.

\bibitem{thiolliere2015hybrid}
Roland Thiolliere, Ewan Dunbar, Gabriel Synnaeve, Maarten Versteegh, and
  Emmanuel Dupoux,
\newblock ``A {H}ybrid {D}ynamic {T}ime {W}arping-deep {N}eural {N}etwork
  {A}rchitecture for {U}nsupervised {A}coustic {M}odeling,''
\newblock in {\em Sixteenth Annual Conference of the International Speech
  Communication Association}. Citeseer, 2015.

\bibitem{badino2015discovering}
Leonardo Badino, Alessio Mereta, and Lorenzo Rosasco,
\newblock ``Discovering {D}iscrete {S}ubword units with {B}inarized
  {A}utoencoders and {H}idden-markov-model {E}ncoders,''
\newblock in {\em Proceedings of Interspeech}, 2015.

\bibitem{baljekar2015using}
Pallavi Baljekar, Sunayana Sitaram, Prasanna~Kumar Muthukumar, and A~Black,
\newblock ``{U}sing {A}rticulatory {F}eatures and {I}nferred {P}honological
  {S}egments in {Z}ero {R}esource {S}peech {P}rocessing,''
\newblock in {\em Proceedings of Interspeech}, 2015.

\end{thebibliography}
\appendices
\section{Time Complexity and Storage Requirement}\label{sec:complexity}

We list time complexity and storage requirement for training the acoustic tokens and performing QbE-STD for a token set of granularity $(m,n)$ under the Multi-granular Paradigm and a lexicon with parameters $(u,v)$ under the Hierarchical Paradigm. 
We ignore the space complexity during the computation of the algorithms, but instead focus on the disk space required to store the results of the algorithms. 
For simplicity we assume every utterance in the spoken archive has the same length $T$, and there are a total of $U$ utterances in the archive.
We list the results in Table \ref{tab:summary} with the corresponding notations in Table \ref{tab:notation}.
Note that there is no distinction between the Multi-granular and Hierarchical Paradigm for most algorithms except for token decoding in Eq. \eqref{eq:3}.


%

 
 
 
\end{document}